\pdfoutput=1

\documentclass[11pt]{article}

\usepackage{acl}

\usepackage{times}
\usepackage{latexsym}

\usepackage[T1]{fontenc}

\usepackage[utf8]{inputenc}
\usepackage{adjustbox}

\usepackage{microtype}

\usepackage{graphicx}
\usepackage{booktabs}
\usepackage{multirow}
\usepackage{array}
\usepackage{tikz}

\usepackage[ruled,vlined,linesnumbered]{algorithm2e}
\SetAlCapNameFnt{\small}
\SetAlCapFnt{\small}
\usepackage[noend]{algpseudocode}
\makeatletter
\def\ALG@special@indent{%
    \ifdim\ALG@thistlm=0pt\relax
        \hskip-\leftmargin
    \else
        \hskip\ALG@thistlm
    \fi
}
\newcommand{\Input}[1]{\item[]\noindent\ALG@special@indent \textbf{Input}\ #1}
\renewcommand{\While}[1]{\item[]\noindent\ALG@special@indent \textbf{while}\ #1 \textbf{do}}
\newcommand{\Statement}[1]{\item[]\noindent\ALG@special@indent #1}
\newcommand{\EndKernel}{\item[]\noindent\ALG@special@indent \textbf{End Kernel}}
\makeatother

\usepackage{amsmath}

\usepackage{amsfonts}

\DeclareRobustCommand\onedot{\futurelet\@let@token\@onedot}
\def\onedot{.}
\def\eg{\emph{e.g}\onedot}

\newcommand{\todo}[1]{}
\newcommand{\lajan}[1]{}
\newcommand{\sung}[1]{}
\newcommand{\anthony}[1]{}
\renewcommand\todo[1]{\textcolor{red}{#1}}
\renewcommand{\lajan}[1]{\textcolor{blue}{#1-LL}}
\renewcommand{\sung}[1]{\textcolor{cyan}{#1-SR}}
\renewcommand\anthony[1]{\textcolor{purple}{#1-AL}}

\newcommand{\ourtitle}{A Picture is Worth a Thousand Words: Language Models Plan from Pixels}

\newcommand{\ours}{VP$^2$\xspace}

\newcommand{\oursfull}{Visual Prompt Planning\xspace}

\newcommand{\fobsembed}{f_\text{obs-enc}}
\newcommand{\cxt}{\mathrm{cxt}}
\newcommand{\concat}{\mathrm{concat}}

\newcommand{\vizalfworld}{VizALF\xspace}
\newcommand{\alfworld}{ALFWorld\xspace}
\newcommand{\vhome}{VirtualHome\xspace}
\newcommand{\gptmed}{GPT2$_\text{med}$\xspace}
\newcommand{\gptxl}{GPT2$_\text{XL}$\xspace}
\newcommand{\gptj}{GPTJ\xspace}
\newcommand{\flan}{FLAN-T5$_\text{xxl}$\xspace}


\usepackage{graphicx}
\usepackage{url}
\usepackage{xspace}
\usepackage{bm}
\usepackage{enumitem}
\usepackage{multirow}
\usepackage{booktabs}
\usepackage{xcolor}

\usepackage[capitalize]{cleveref}
\crefname{section}{Sec.}{Secs.}
\Crefname{section}{Section}{Sections}
\Crefname{table}{Table}{Tables}
\crefname{table}{Tab.}{Tabs.}

\usepackage{cleveref}
\usepackage{pgfplots} %
\pgfplotsset{width=6.5cm, compat=1.6}

\usepackage{caption}
\usepackage{subcaption}

\usepackage{microtype}

\title{\ourtitle}

\author{Anthony Liu\thanks{$\,\,\,$Work done during an internship with LG AI Research. Correspondence to \tt{anthliu@umich.edu}.}\hspace{3pt}, Lajanugen Logeswaran, Sungryull Sohn, Honglak Lee \\ LG AI Research}

\begin{document}
\maketitle

\begin{abstract}
Planning is an important capability of artificial agents that perform long-horizon tasks in real-world environments.
In this work, we explore the use of pre-trained language models (PLMs) to reason about plan sequences from text instructions in embodied visual environments.
Prior PLM based approaches for planning either assume observations are available in the form of text (e.g., provided by a captioning model), reason about plans from the instruction alone, or incorporate information about the visual environment in limited ways (such as a pre-trained affordance function). %
In contrast, we show that PLMs can accurately plan even when observations are directly encoded as input prompts for the PLM.
We show that this simple approach outperforms prior approaches in experiments on the ALFWorld and VirtualHome benchmarks.
\end{abstract}

\section{Introduction}

The ability to reason about plans is critical for performing long-horizon tasks \citep{erol1996hierarchical, sohn2018hierarchical, sharma-etal-2022-skill}, compositional generalization \citep{corona-etal-2021-modular} and generalization to unseen tasks and environments \citep{shridhar2020alfred}.
Consider a simple long-horizon planning scenario where a robot is tasked with preparing a meal and serving it on the table. 
This presents a non-trivial planning problem since the agent needs to understand the sequence of operations required to perform the task and search for the relevant objects in the unfamiliar environment by interacting with various objects. %

Large language models have been recently shown to possess commonsense knowledge about the world such as object affordances and physical dynamics \citep{ouyang2022training,chowdhery2022palm}.
Early approaches considered text based environments and fine-tuned PLMs to predict actions given the history of past observations and actions \citep{jansen-2020-visually,micheli-fleuret-2021-language,yao-etal-2020-keep}.
Recent work has used this ability to reason about plans from text instructions in simulated household environments with simplifying assumptions such as text-only environment observations or feedback \citep{huang2022language,ahn2022can,li2022pre,logeswaran-etal-2022-shot}.

We focus on \emph{visually grounded planning} with PLMs --- the ability to adapt plans based on interaction and visual feedback from the environment.
While PLMs have strong planning commonsense priors, predictions from a PLM may not be directly realizable in the environment since the observation and action spaces are unknown.
This requires \emph{grounding} the PLM in the environment and adapting it to observe visual feedback, which is highly non-trivial.
Some prior works assume the availability of a pre-trained affordance function \citep{ahn2022can} or a success detector \citep{mirchandani2021ella}.
Notably, SayCan \citep{ahn2022can} completely decouples the PLM from observation information by selecting actions that have both high affordability (through a pre-trained affordance model) and high PLM likelihood.
Although this partially addresses the grounding problem, the use of visual feedback for action affordance alone is limited.
Often an agent must choose one of many affordable actions using information from observations.
For example, a driving agent should re-navigate and possibly turn around when encountering a ``road closed'' sign, but both turning around and driving forward are indistinguishable to SayCan because they are both affordable and the PLM is blind to observations.

Another workaround explored in prior work is translating the information in the visual observations to text using a pre-trained captioning system \citep{shridhar2021alfworld,huang2022language}.
However, it can be difficult to faithfully describe an image in words and information is lost in this inherently noisy process, which limits the information available to the planner.

Recent work shows that PLMs can be adapted for various natural language tasks by inserting tunable embeddings or soft prompts at the input of the PLM (also called prompt tuning or prefix tuning)~\citep{li-liang-2021-prefix,lester-etal-2021-power}.
This approach also extends to multi-modal understanding tasks such as image captioning \citep{mokady2021clipcap} and VQA \citep{tsimpoukelli2021multimodal} where images are encoded as soft prompts and finetuned for the target task.
Transformer based architectures have also been successfully applied to offline Reinforcement Learning in recent work \citep{chen2021decision,janner2021offline,li2022pre,reid2022can}.

Taking inspiration from these works, we propose the simple approach of embedding visual observations (`visual prompts') and \textit{directly inserting them as PLM input embeddings}.
The visual encoder and PLM are jointly trained for the target task, an approach we call \textbf{\oursfull}~(\ours).
By teaching the PLM to use observations for planning in an end to end manner, we remove the dependency on external data such as captions and affordability information that was used in prior work.
We show that this simple approach performs better than prior PLM-based planning approaches on two embodied planning benchmarks based on ALFWorld~\citep{shridhar2021alfworld} and Virtualhome~\cite{puig2018virtualhome}.

\section{Preliminary: Prompt Tuning}
\label{sec:prelim}

Given a sequence of tokens $x_1,\ldots,x_t$, an auto-regressive language model predicts a probability distribution over the next token $p_\text{LM}(\cdot|x_1,\ldots,x_t)$.
While a PLM expects to see natural language tokens in its context, the model can be extended to process a sequence of input embeddings. %
The input layer of a PLM converts tokens $x_1,\ldots,x_t$ into token embeddings $e_1,\ldots,e_t$ which are passed on to subsequent layers.
Soft prompt tuning introduces additional tunable embeddings $p_1,\ldots,p_k$ in the input layer $p_\text{LM}(\cdot|e_1,\ldots,e_t,p_1,\ldots,p_k)$\footnote{In an abuse of notation, we will use $p_\text{LM}$ with token inputs or embedding inputs interchangeably.} which can be optimized with respect to a target training objective using gradient descent.
In addition, for cross modal reasoning tasks the soft embeddings $p_i$ can be a function of data-modalities other than text such as images (e.g., $p_i = f_i(v;\theta)$ where $v$ is an image and $\theta$ are trainable parameters).
We use this approach to augment the language model context with visual observations.
\section{Approach}

\paragraph{Problem Setting.}
We assume a goal-based MDP setting, parameterized by $\mathcal{M} = \left(\mathcal{S}, \mathcal{A}, \mathcal{G}, P, R_\mathcal{G}\right)$: a state space $\mathcal{S}$, action space $\mathcal{A}$, a goal space $\mathcal{G}$, transition probabilities $P$, and reward $R_\mathcal{G}$. %
The planner is given $N$ expert demonstrations 
$\mathcal{D} = \{ (g^{(i)}, o_0^{(i)}, a_0^{(i)}, o_1^{(i)}, a_1^{(i)}, \dots, o_T^{(i)}, a_T^{(i)}) \}_{i=1}^N$
where goals $g^{(i)}\in \mathcal{G}$ and actions $a^{(i)} \in \mathcal{A}$ are available as text and observations $o^{(i)} \in \mathbb{R}^{H \times W \times C}$ are images of size $H\times W\times C$.
Further, we do not assume the list of possible actions available to the agent is known, or any pretrained admissibility or affordance function is known. 
Given goal description $g$, past actions $a_1, \ldots, a_{t-1}$ and observations $o_1, \ldots, o_t$, we seek to build a policy $\pi$ which models the next action probability $\pi(a_t|g,a_{1\cdots t-1},o_{1\cdots t})$.

\paragraph{\oursfull.}
If goal description, actions and observations are available in the form of discrete token sequences, predicting the next action is similar to a language modeling task and a PLM can be fine-tuned for next action prediction: maximize $\log p_\text{LM} ( a_t \mid \cxt_t)$, where  
$\cxt_t = \concat\left(g, o_1, a_1, o_2, a_2, \dots, a_{t-1}, o_t\right)$.
However, observations may not be available in the form of text or discrete tokens in practice and we attempt to tackle this scenario.

As we discuss in \Cref{sec:prelim}, PLMs are capable of processing a sequence of embeddings (which may not necessarily correspond to actual text tokens).
The context can be re-written in terms of embedding sequences as 
$\cxt_t = \concat\left(g^e, o^e_1, a^e_1, o^e_2, a^e_2, \dots, a^e_{t-1}, o^e_t\right)$\footnote{Note that each of $g^e, o^e_i, a^e_i$ are embedding sequences and the concat operation concatenates these sequences.} 
where $g^e, o_i^e, a_i^e$ respectively represent \emph{embedding sequences} corresponding to the goal description, observations and actions.
We assume that the goal description and actions are available in the form of text and $g^e, a_i^e$ can be obtained as the corresponding token embedding sequences.

\paragraph{Observation Encoder.}
To obtain observation embeddings $o_i^e$, we propose to learn an observation encoder $f: \mathbb{R}^{H \times W \times C} \to \mathbb{R}^{m \times E}$. $f$ maps visual observations $o$ to a sequence of $m$ embeddings each of size $E$, where $m$ is a hyperparameter and $E$ is the PLM's embedding dimensionality. %
In our experiments we consider an observation encoder of the form shown in \Cref{eq:obsenc} where $f_\text{pretrained}$ is a pre-trained visual encoder and $f_\text{FFN}$ is a feedforward network.
\begin{equation}
\fobsembed(o) = f_\text{FFN}(f_\text{pretrained}(o))
\label{eq:obsenc}
\end{equation}

\paragraph{Training Objective.}
We learn to model the next action given previous actions, observations, and the goal. %
Similar to prior approaches~\cite{micheli-fleuret-2021-language, huang2022language}, we define the loss as in \Cref{eq:trainobj},
where $\cxt^{(i)}_t$ is a concatenation of goal, action and observation embeddings as previously described.
\begin{equation}
\mathcal{L}_\mathcal{D} = - \frac{1}{N} \sum_{i, t} \log p_\text{LM} (a^{(i)}_t \mid \cxt^{(i)}_t)
\label{eq:trainobj}
\end{equation}

\section{Experiments}

\subsection{Environments}

We experiment with embodied agent tasks that involve navigating and manipulating objects in a simulated household environment.
The agent acts by feeding text commands to a low-level controller that executes various pretrained skills (such as \textit{go to cabinet} or \textit{take apple from cabinet}).

\paragraph{\vizalfworld.}
This environment is based on \alfworld~\citep{shridhar2021alfworld} and contains 6 types of tasks that are compositional and contain multiple subgoals that must be completed.
In contrast to ALFWorld which is a purely text based environment, we consider the same set of tasks but with \textbf{only} visual observations from the AI2-Thor simulator \citep{ai2thor}.
We used the training and evaluation task split provided in ALFWorld which consists of 4620 training tasks, 187 in distribution evaluation tasks, and 192 out of distribution evaluation tasks.
However, we found 64/187 and 52/190 of the ID and OD evaluation tasks respectively were impossible to complete, due to errors in the ALFWorld low level action implementations. So, we normalized the success rate of all agents by the oracle agent's success rate.

\paragraph{VirtualHome.}
We experiment with tasks from LID \citep{li2022pre} which are based on the VirtualHome simulator \citep{puig2018virtualhome}.
Each task is specified using a set of goal conditions that must be met at the end of the episode (e.g., There must be two pancakes in the fridge and the stove must be turned on).
We use the in distibution and novel scene splits from LID. 2000 in distribution tasks were used for training and 200 novel scene tasks were used for evaluation.

\subsection{Models}

\begin{table*}[!ht]
    \centering
    \begin{tabular}{p{20em} >{\centering\arraybackslash}p{3em} >{\centering\arraybackslash}p{3em} >{\centering\arraybackslash}p{3em} >{\centering\arraybackslash}p{3em}}
        \toprule
        \multicolumn{1}{c}{\multirow{2}{*}[-0.2em]{Approach}} & \multicolumn{2}{c}{\vizalfworld} & \multicolumn{2}{c}{VirtualHome}\\
        \cmidrule(lr){2-3} \cmidrule(lr){4-5}
        & ID & OD & ID & OD \\
        \midrule
        Ignore                  \citep{jansen-2020-visually}   & 35.4 & 22.4 & 6.0 & 1.5 \\
        Captions                \citep{shridhar2021alfworld}   & 53.2 & 20.3 & 7.0 & 2.5 \\
        SayCan \citep{ahn2022can} & & & & \\
        \hspace{0.5em} - \gptmed~{\small(\emph{Finetuned PLM, Trained affordance})} & 36.6 & 19.3 & 7.0 & 1.0 \\% 24.1 & 14.1 & . & .
        \hspace{0.5em} - \gptmed~{\small(\emph{Finetuned PLM, Oracle affordance})}  & 51.2 & 27.8 & 13.6 & 5.0 \\
        \hspace{0.5em} - \flan~{\small(\emph{Few-shot PLM, Oracle affordance})}  & 40.6 & 26.1 & 3.9 & 0.0 \\% 26.7 & 19.0 & 3.9 & 0.00
        \ours (ours)                & \textbf{55.3} & \textbf{27.8} & \textbf{20.6} & \textbf{7.5} \\
        \bottomrule
    \end{tabular}
    \caption{Success rates of approaches on \vizalfworld and \vhome (VH) benchmarks. We present the average success rate for in distribution (ID) and out of distribution (OD) tasks.}
    \label{tab:vizalf-results}
\end{table*}

We use the \gptmed language model (355M parameters) in all our experiments.
We consider the following baselines for comparison.
    \paragraph{Ignore.} A simple baseline inspired by \citet{jansen-2020-visually} that ignores the visual observations and predicts the entire text action sequence only from the goal text description. This baseline is finetuned with the same objective as in \Cref{eq:trainobj} but without observations in the context.
    \paragraph{Captions.} Instead of feeding visual observations to the planner language model, we use text captions predicted by a captioning model as a proxy \citep{shridhar2021alfworld}. We train a ClipCap \citep{mokady2021clipcap} model on ground-truth captions from the respective environment's training demonstrations and use them for captioning. The captioning model is trained on 70k and 60k captions on \vizalfworld and \vhome respectively.
    \paragraph{SayCan.}
    The SayCan~\citep{ahn2022can} architecture has two components: a) A PLM that ranks actions and b) An affordance function that predicts what actions are affordable from a given state.
    SayCan evaluates a given action by combining its likelihood under the PLM (ignoring visual observations) with its affordance score as shown in \Cref{eq:saycan}.
    \begin{align}
        \mathrm{Score}(a_t) = p_\mathrm{LM}(a_t | g, a_{t-1}, \dots, a_1) \cdot p_\mathrm{aff}(a_t | o_t)
        \label{eq:saycan}
    \end{align}
    We consider different design choices for each of these components.
    We consider two variations for the PLM: 
    1) A frozen \flan (11B) model that is few-shot prompted similar to the original work.
    2) A \gptmed model fine-tuned for next action prediction (details in the appendix).

    We also consider two variations for the affordance model: 
    1) An \emph{oracle affordance} function which assumes knowledge about ground-truth affordable actions. %
    2) A \emph{trained affordance} function trained to predict whether an action is affordable from a given visual observation using supervised learning on training demonstrations annotated with affordance information.
        In both versions, we use a PLM to predict action sequences from the goal text and select an action based on the SayCan score in \Cref{eq:saycan}.

\subsection{Results}

Table~\ref{tab:vizalf-results} compares the performance of our method against baselines.
Our simple approach (\ours) performs better than all baselines, despite not using external data (caption and affordance information).
\ours benefits from direct coupling between the planner language model and environment observations.

The Ignore baseline performs worse compared to other methods that make use of observations.
However, on out of distribution tasks it suffers less from domain-shift compared to some of the other grounding baselines such as Captions.
The Captions baseline performs better than Ignore, but suffers from information loss in the captioning process.

In comparison, SayCan with oracle affordance is comparable to or better than Ignore and Captions (slightly worse than Captions on \vizalfworld ID) in spite of incorporating observation information only through the affordance function (both \flan and \gptmed).
SayCan using the trained affordance only performs slightly better than Ignore on in distribution, and similarly suffers from domain-shift on out of distribution tasks.
\flan performs well on \vizalfworld despite no training.
However, it performs poorly on \vhome due to demonstration trajectories in \vhome being relatively long, which inhibits SayCan from using many examples for few-shot prompting.

\subsection{Ablations}

\begin{table}[ht]
    \centering
    \begin{tabular}{llrr}
        \toprule
        \multirow{2}{*}[-0.2em]{Ablation} & & \multicolumn{2}{c}{\vizalfworld} \\
        \cmidrule{3-4}
        & & ID & OD \\
        \midrule
        & \ours &                                    55.3  & 27.8  \\
        \midrule
        visual enc.\
        & resnet50 &                  30.1  & 12.9  \\
        \midrule
        prompt & CLIPCap &            50.5  & 15.9  \\
        \midrule
        \multirow{3}{*}{base LM}
        & \gptmed Frozen\ &                            48.0  & 16.5  \\
        & \gptxl Frozen\ &                             48.0  & 16.7  \\
        & \gptj Frozen\ &                              50.0  & 15.8  \\
        \bottomrule
    \end{tabular}
    \caption{Select ablations on the \vizalfworld benchmark, with average success rate for in distribution (ID) and out of distribution (OD) tasks. We tested our approach \ours with and without different components. A full ablation table is included in the appendix (Table~\ref{tab:vizalf-ablations-full}).}
    \label{tab:vizalf-ablations}
\end{table}

Table~\ref{tab:vizalf-ablations} presents ablations we perform to identify the importance of each component of our approach.

\paragraph{Visual Encoder.}
Replacing the CLIP visual encoder with a Resnet50 \citep{he2016deep} significantly degrades the performance.
This suggests that the image-text alignment pre-training of CLIP helps produce observation features that are more easily interpreted by the language model.
In contrast to prior methods that consider auxiliary alignment objectives to match the distribution of inputs \citep{reid2022can}, it could be more beneficial to use powerful encoders such as CLIP pre-trained for image-text alignment.

\paragraph{Pretrained Prompt Model.}
We also test how using pre-trained visual prompts can affect performance.
We used visual prompts pre-trained with the CLIPCap captioning objective on the Conceptual Captions dataset~\citep{conceptual}.
However, using this pre-trained visual prompt hurts the success rate.
We hypothesize that the knowledge aquired by the LM during captioning isn't directly useful for action prediction.

\paragraph{Prompt Tuning on frozen language models.}
Finally, we consider an ablation where the language model backbone is held fixed and only visual prompts and task prompt tokens are tuned (similar to Frozen \citep{tsimpoukelli2021multimodal}).
We found that freezing the language model generally performs worse than fine-tuning it. 
The drop in performance can be attributed to the limited influence of prompt tokens alone in controlling the language model's behavior.

\paragraph{Effect of LM pre-training.}
\begin{figure}[!t]
\centering
\includegraphics[trim=5 10 0 20,width=0.48\textwidth]{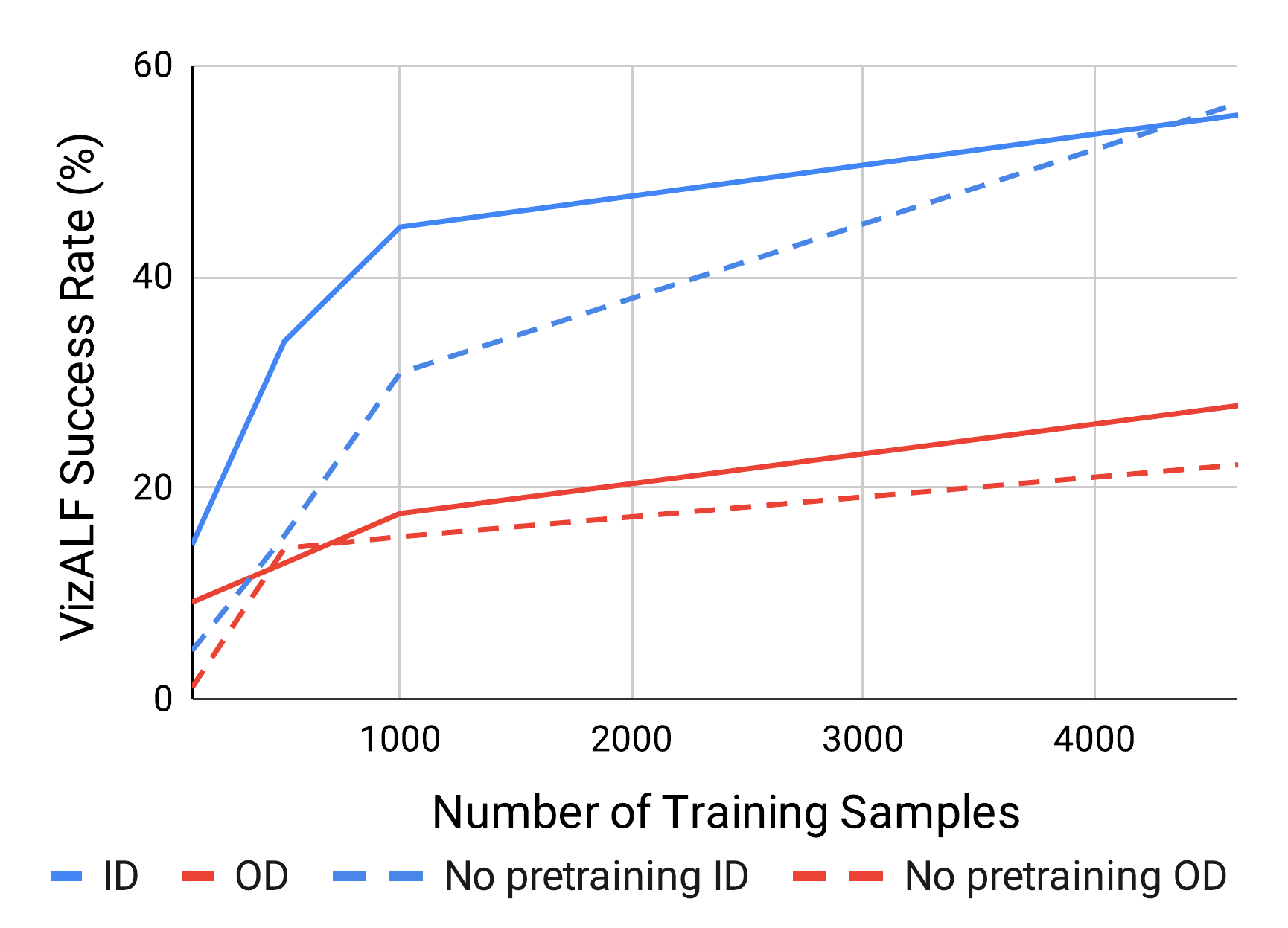}
\caption{Ablation showing the learning efficiency (a) with and without language model pre-training and (b) varying the number of training demonstrations.}
\label{fig:samples-pt}
\end{figure}
We examine the benefit of LM pre-training by training a model from scratch on varying amounts of training data in Figure~\ref{fig:samples-pt}.
We find that the pre-trained model converges faster and is more sample efficient compared to the model trained from scratch.
This confirms findings from prior work about how language model pre-training benefits learning sequential decision making tasks \citep{reid2022can}.

\section{Conclusion}
We present a simple approach for planning from pixels building on the planning commonsense knowledge acquired by large language models.
Compared to prior work, which indirectly incorporates observation information by captions or affordance, our approach is simpler, does not use external data, and benefits from directly coupling the planner language model and environment observations.
Experimentally, we showed our approach performs better than these prior methods on two embodied agent benchmarks.

\bibliography{anthology,refs}
\bibliographystyle{acl_natbib}

\appendix
\appendix
\section{Appendix}

\subsection{Experimental Setup Details}
\paragraph{Visual Observations, Text Actions.}
\vizalfworld uses the same action set defined in \alfworld with the following difference. 
\alfworld is a text-based environment and it references multiple instances of an object type with object identifiers (cabinet 1, cabinet 2, etc.).
These object identifiers are no longer meaningful in the visual setting as the grounding of cabinet 1 to the physical cabinet is unknown to the agent.
We thus removed these numeric identifiers in the action and ground-truth text observation (\eg, ``cabinet 1'' becomes ``cabinet'').
The set of valid actions and objects is defined in~\citep{shridhar2021alfworld}.
We use the same procedure for removing object numeric identifiers from \vhome. 

\paragraph{Ground Truth Captions.}
Ground truth captions are needed to train the captions baseline and needed for the captioning auxiliary task. In \vizalfworld, we use the ground truth captions provided in ALFWorld~\citep{shridhar2021alfworld}, which are generated from visible objects and a pre-defined template.
We define ground truth captions for \vhome using a similar template on the environment's list of interactable objects for each observation.

\paragraph{Ground Truth Affordability.}
Ground truth labels for affordability are needed to train the SayCan trained affordance model. As the trained affordance is a binary classifier, we create a training set of observation-action-affordability pairs. To do this, for each observation, we sampled all affordable actions, and a subset of non-affordable actions (as there are vastly more non-affordable actions in each state).
We tested two methods for sampling non-affordable actions: 1) We used a hand-coded heuristic that selects non-affordable actions based on actions that are likely to be predicted by the LM but are not affordable.
2) We use actions that are predicted with high likelihood from the Ignore baseline but are not affordable.
We did not find a significant difference in non-affordable examples on performance.

\subsection{SayCan Baseline Details}
In the original SayCan~\citep{ahn2022can}, the PLM gives a score for every possible action at each timestep.
When the oracle action affordances are known (around 15 actions affordable) at a given time,
we can evaluate the PLM score on all affordable actions.
However, when oracle affordances are not available, in our trained affordance setting, each step has \textit{thousands} of possible actions (actions are a combinatorial product on action type and objects). Evaluating the LM on each action is infeasible.
Instead, we use a beam search on the PLM to generate the top-$k$ most likely actions, rank
these likely actions using the SayCan score. Specifically, for a given observation $o$, predict the next action $a$ by:
\begin{enumerate}[leftmargin=*,itemsep=0em]
\item Sample top-$k$ actions $a_1, \dots, a_k$ from PLM.
\item Compute
    \begin{align*}
        \mathrm{Score}(a_t) = p_\mathrm{LM}(a_t | g, a_{t-1}, \dots a_1) \cdot p_\mathrm{aff}(a_t | o_t)
        \label{eq:saycan}
    \end{align*}
\item Return $\mathrm{argmax}(\text{Score}(a))$
\end{enumerate}

In addition, the original SayCan does not perform any finetuning on the PLM. 
However, we found that SayCan improved in our environment tasks when finetuning a smaller model.

\subsection{Code Implementation Details}

Our approaches and baselines were implemented using the Huggingface Transformers Python library~\cite{wolf2019huggingface}.
Code for the \vizalfworld environment was modified from the ALFWorld codebase~\cite{shridhar2021alfworld}.
Code for the \vhome environment was modified from the LID codebase~\cite{li2022pre}.

\subsection{Computational Budget}
Experiments were conducted on A100 GPUs. Most approaches used a \gptmed as a base LM, totalling 400M parameters. Prompt tuning ablations with larger frozen models had 1.5B and 6B parameters in total.
For the \flan-SayCan experiments (using an 11B model) we did no training.

\subsection{Hyperparameters}

In general we used similar hyperparameters for the PLM and visual prompts across approaches, but adjusted the epochs to account for learning visual prompts vs. learning text only and number of training samples.

\begin{table}[ht!]
    \centering
    \begin{tabular}{lr}
        \toprule
        Hyperparameter & Value \\
        \midrule
        Epochs & 50 \\
        Batch Size & 8 \\
        Seed & \{0, 1\} \\
        Grad Accum.\ Steps & 1 \\
        LM & \gptmed \\
        LM learning rate & 5e-5 \\
        LM weight decay & 1e-3 \\
        Gradient clipping & None \\
        Weight decay & 0.01 \\
        VP learning rate & 1e-2 \\
        VP size & 10 \\
        VP visual encoder & CLIP ViT-B/32\\
        VP arch.\ & 2-hidden-layer MLP \\
        Max context tokens & 1000 \\
        \bottomrule
    \end{tabular}
    \caption{Hyperparameters for \ours in \vizalfworld and \vhome.}
    \label{tab:vp2hyperparameters}
\end{table}
\paragraph{\ours.} We show relevant hyperparameters for \ours in Table~\ref{tab:vp2hyperparameters}.

\paragraph{Ignore Baseline.}
We used the same hyperparameters as in Table~\ref{tab:vp2hyperparameters} but remove visual prompt parameters.

\paragraph{Caption Baseline.}
To train a captioning model, we use the same hyperparameters as Table~\ref{tab:vp2hyperparameters} but set \{epochs = 10\}.
We train a separate PLM for the action prediction model (which uses predicted captions as input) using the same hyperparameters as Table~\ref{tab:vp2hyperparameters} but set \{epochs = 20\}.
For the context of the captions, we add goal text to encourage the PLM to produce captions: 
``Your task is to: caption the following observation''.
We also tested finetuning a CLIPCap captioning model for this baseline, but found this decreased captioning performance.

\paragraph{SayCan Baseline.}
For the frozen \flan-SayCan model, we performed prompt engineering similar to~\citet{ahn2022can}. For every task, we sample $k$ examples for few shot prompting (as many can fit in the context length) in the following format:

{\small
\begin{verbatim}
Here are some step by step instructions 
for example tasks.
Example: <g1>. 1. <a1> 2. [...] [...]
Example: <gk>. 1. <a1> 2. [...]
Give step by step instructions for the 
following task. 
<goal>
\end{verbatim}
}

To retrieve examples, we sampled $k$ samples from the training dataset that are similar to the current goal.
For \vizalfworld, we took samples that are the same task type (pick-place, heat, etc.\ ) and
for \vhome, we took samples where the goals were most similar according to a simple bag of words heuristic.
We found the best value for $k = 30$ from a grid search on $k \in \{1, 5, 10, 15, 30\}$.

For the fine-tuned action prediction model \gptmed-SayCan, we use the same hyperparameters as Table~\ref{tab:vp2hyperparameters} but set \{epochs = 20\}.
For the trained affordance model, we use the same hyperparameters as
Table~\ref{tab:vp2hyperparameters} but set \{epochs = 2\}.
For the context of the trained affordance, we add goal text to encourage the PLM to predict affordance: 
``Your task is to: predict whether the following action is valid.''
The PLM must either output the token ``valid'' or ``invalid'', given the observations and action contexts.

\subsection{Full Ablations}

\begin{table}[ht!]
    \centering
    \begin{tabular}{llrr}
        \toprule
        \multirow{2}{*}[-0.2em]{Ablation} & & \multicolumn{2}{c}{\vizalfworld} \\
        \cmidrule{3-4}
        & & ID & OD \\
        \midrule
        & \ours &                                    55.3  & 27.8  \\
        \midrule
        visual enc.\
        & resnet50 &                  30.1  & 12.9  \\
        \midrule
        prompt & CLIPCap &            50.5  & 15.9  \\
        \midrule
        \multirow{3}{*}{base LM}
        & \gptmed Frozen &                            48.0  & 16.5  \\
        & \gptxl Frozen &                             48.0  & 16.7  \\
        & \gptj Frozen &                              50.0  & 15.8  \\
        \midrule
        \multirow{3}{*}{prompt size}
        & 1 &                      48.8  & 14.9  \\
        & 5 &                      54.4  & 17.1  \\
        & 20 &                    54.8  & 20.0  \\
        \midrule
        \multirow{2}{*}{aux.\ task}
        & inv-dyn.\ &                           58.8  & 20.7  \\
        & captions &                              58.5  & 21.4   \\
        & goal-pred.\ &                               52.9 & 17.8 \\
        \midrule
        \multirow{3}{*}{samples}
        & 100 &                      14.6  & 9.2  \\
        & 500 &                      33.9  & 12.9  \\
        & 1000 &                     44.7  & 17.6  \\
        \midrule
        \multirow{4}{*}{\parbox{1.8cm}{samples, no pre-train}}
        & 100 &         4.6  & 1.1  \\
        & 500 &         15.5  & 14.3  \\
        & 1000 &        30.9  & 15.4  \\
        & 4620 (all) &         56.4  & 22.2  \\
        \bottomrule
    \end{tabular}
    \caption{All ablations on the \vizalfworld benchmark, with average success rate for in distribution (ID) and out of distribution (OD) tasks. We tested our approach \ours with various components added or removed.}
    \label{tab:vizalf-ablations-full}
\end{table}

We present more detailed ablations in 
\Cref{tab:vizalf-ablations-full} presents additional ablations for prompt size, training samples, and LM pretraining (more comprehensive version of  \Cref{tab:vizalf-ablations} in the main text).

\begin{figure*}[!t]
\centering
\includegraphics[width=0.9\textwidth]{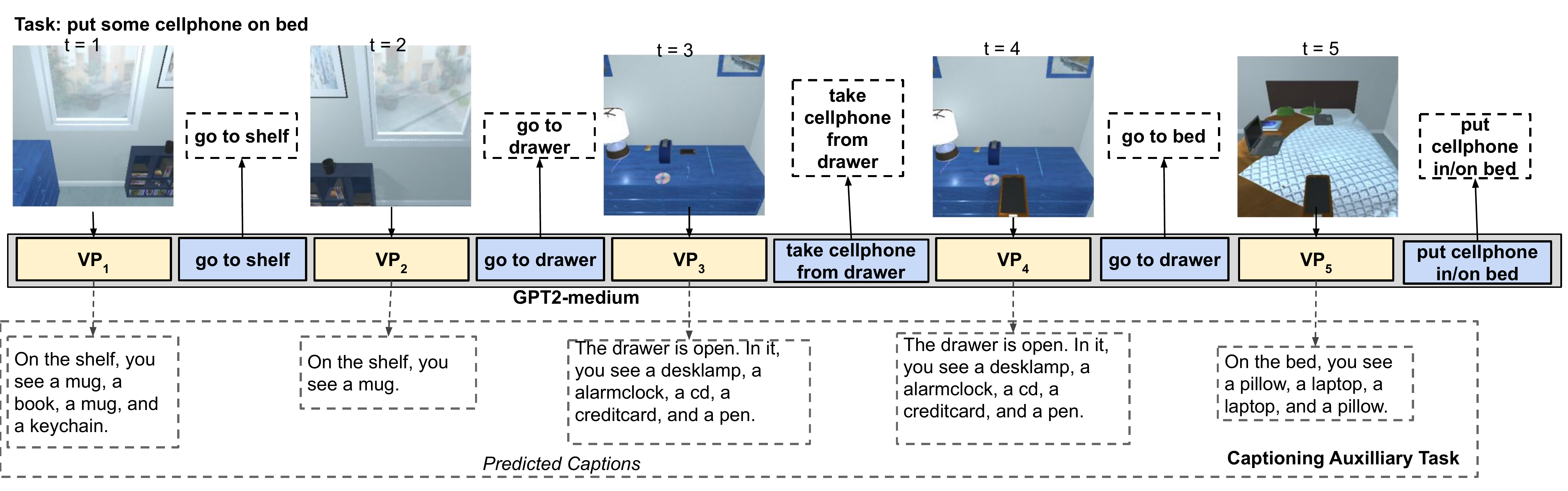}
\caption{Example of \ours with the captioning auxiliary task in \vizalfworld. Each observation is encoded into a visual prompt VP$_t$ and used to predict the next action.
The LM is also trained to predict a caption for each VP$_t$.}
\label{fig:aug-arch}
\end{figure*}

\paragraph{Prompt Size.}
We tested visual prompt sizes \{1, 5, 10, 20\}, where 10 is the prompt size used in \ours.
Lowering the prompt size to 1 and 5 lowers the success rate in \vizalfworld, as less visual information can be contained in each prompt.
However, raising the prompt size to 20 may allow more information in a visual prompt, but also causes the LM's context to increase. This can harm the LM as the context needs to be trimmed to fit within the limited context window.
From Table~\ref{tab:vizalf-ablations-full}, this seems to harm the success rate.

\subsection{Auxiliary Tasks}
In \ours, the LM is only trained on the action prediction loss function $\mathcal{L}_\mathcal{D}$.
We hypothesized that auxiliary tasks that train visual prompts in additional ways can help ground and improve the visual prompts for planning.
To do this, we trained the LM concurrently on $\mathcal{L}_\mathcal{D}$ and a loss derived for each auxiliary task.
1) Inverse Dynamics (inv-dyn.\ ). The LM must predict the action that is executed between two observations:
$\mathcal{L}_\text{inv-dyn} = -\frac{1}{N} \sum_{i,t} \log p_\text{LM}(a_t^{(i)} | o_t^{(i)}, o_{t+1}^{(i)})$.
2) Captions. The LM must predict the ground truth caption text for each observation.
This is the same training objective used in CLIPCap~\citep{mokady2021clipcap}:
$\mathcal{L}_\text{cap} = -\frac{1}{N} \sum_{i,t} \log p_\text{LM}(\text{caption}_t^{(i)} | o_t^{(i)})$.
3) Goal Prediction (goal-pred). The LM must predict the goal text given action-observation context.
$\mathcal{L}_\text{goal-pred} = -\frac{1}{N} \sum_{i} \log p_\text{LM}(g^{(i)} | o_1^{(i)}, a_1^{(i)}, \dots, o_T^{(i)}, a_T^{(i)})$.

\paragraph{Results.}
We show an example of the captioning auxiliary task in Figure~\ref{fig:aug-arch}.
We find auxiliary tasks (inv-dyn.\ and captions) can help ground visual prompts and improve success rate for in distribution tasks. However they also cause \ours to overfit and perform worse on out of distribution tasks.
The last auxiliary task goal pred.\ seems to decrease performance in both ID and OD.

\paragraph{Hyperparameters.}
For auxiliary tasks, we add a tunable task embedding to each context, where each task (action prediction vs.\ auxiliary task) has a separate task embedding. This embedding helps the LM to learn multiple tasks. We used a task embedding length = 10.
To compute the loss, we add a weight parameter to control the loss between action prediction and auxiliary task: $\mathcal{L} = \mathcal{L}_\mathcal{D} + \alpha \mathcal{L}_\text{aux}$.
In our experiments, we tested $\alpha = \{0.1, 1.0\}$ and found $\alpha = 0.1$ works the best.
Otherwise, we used the same hyperparameters as Table~\ref{tab:vp2hyperparameters}.

\end{document}